\documentclass[sigconf]{acmart}
\usepackage{fancyhdr}
\usepackage{amsfonts}
\usepackage{amsmath}
\usepackage{subfigure}
\usepackage{balance}
\usepackage{url}
\AtBeginDocument{%
  \providecommand\BibTeX{{%
    \normalfont B\kern-0.5em{\scshape i\kern-0.25em b}\kern-0.8em\TeX}}}

\copyrightyear{2021} 
\acmYear{2021} 
\setcopyright{acmcopyright}\acmConference[MM '21]{Proceedings of the 29th ACM
International Conference on Multimedia}{October 20--24, 2021}{Virtual Event, China}
\acmBooktitle{Proceedings of the 29th ACM International Conference on Multimedia
(MM '21), October 20--24, 2021, Virtual Event, China}
\acmPrice{15.00}
\acmDOI{10.1145/3474085.3475708}
\acmISBN{978-1-4503-8651-7/21/10}

\begin{document}
\fancyhead{}
\title{RCNet: Reverse Feature Pyramid and Cross-scale Shift Network for Object Detection}

\author{Zhuofan Zong$^{1}$,\quad Qianggang Cao$^{1}$,\quad Biao Leng$^{1\dagger}$}

\makeatletter
\def\authornotetext#1{
\if@ACM@anonymous\else
    \g@addto@macro\@authornotes{
    \stepcounter{footnote}\footnotetext{#1}}
\fi}
\makeatother
\authornotetext{Corresponding author.}

\affiliation{
 \institution{\textsuperscript{\rm 1}School of Computer Science and Engineering, Beihang University}
 }
\email{{zongzhuofan,  xueyeeeh}@gmail.com, lengbiao@buaa.edu.cn}

\def\authors{Zhuofan Zong, Qianggang Cao, Biao Leng}

\renewcommand{\shortauthors}{Zhuofan Zong et al.}

\renewcommand{\shortauthors}{Zhuofan Zong et al.}


\begin{abstract}

Feature pyramid networks (FPN) are widely exploited for multi-scale feature fusion in existing advanced object detection frameworks. Numerous previous works have developed various structures for bidirectional feature fusion, all of which are shown to improve the detection performance effectively. We observe that these complicated network structures require feature pyramids to be stacked in a fixed order, which introduces longer pipelines and reduces the inference speed. Moreover, semantics from non-adjacent levels are diluted in the feature pyramid since only features at adjacent pyramid levels are merged by the local fusion operation in a sequence manner. To address these issues, we propose a novel architecture named RCNet, which consists of Reverse Feature Pyramid (RevFP) and Cross-scale Shift Network (CSN). RevFP utilizes local bidirectional feature fusion to simplify the bidirectional pyramid inference pipeline. CSN directly propagates representations to both adjacent and non-adjacent levels to enable multi-scale features more correlative. Extensive experiments on the MS COCO dataset demonstrate RCNet can consistently bring significant improvements over both one-stage and two-stage detectors with subtle extra computational overhead. In particular, RetinaNet is boosted to 40.2 AP, which is 3.7 points higher than baseline, by replacing FPN with our proposed model. On COCO \textit{test-dev}, RCNet can achieve very competitive performance with a single-model single-scale 50.5 AP. Codes will be made available.

\end{abstract}




\maketitle

\begin{figure}[h]
  \centering
  \includegraphics[width=\linewidth]{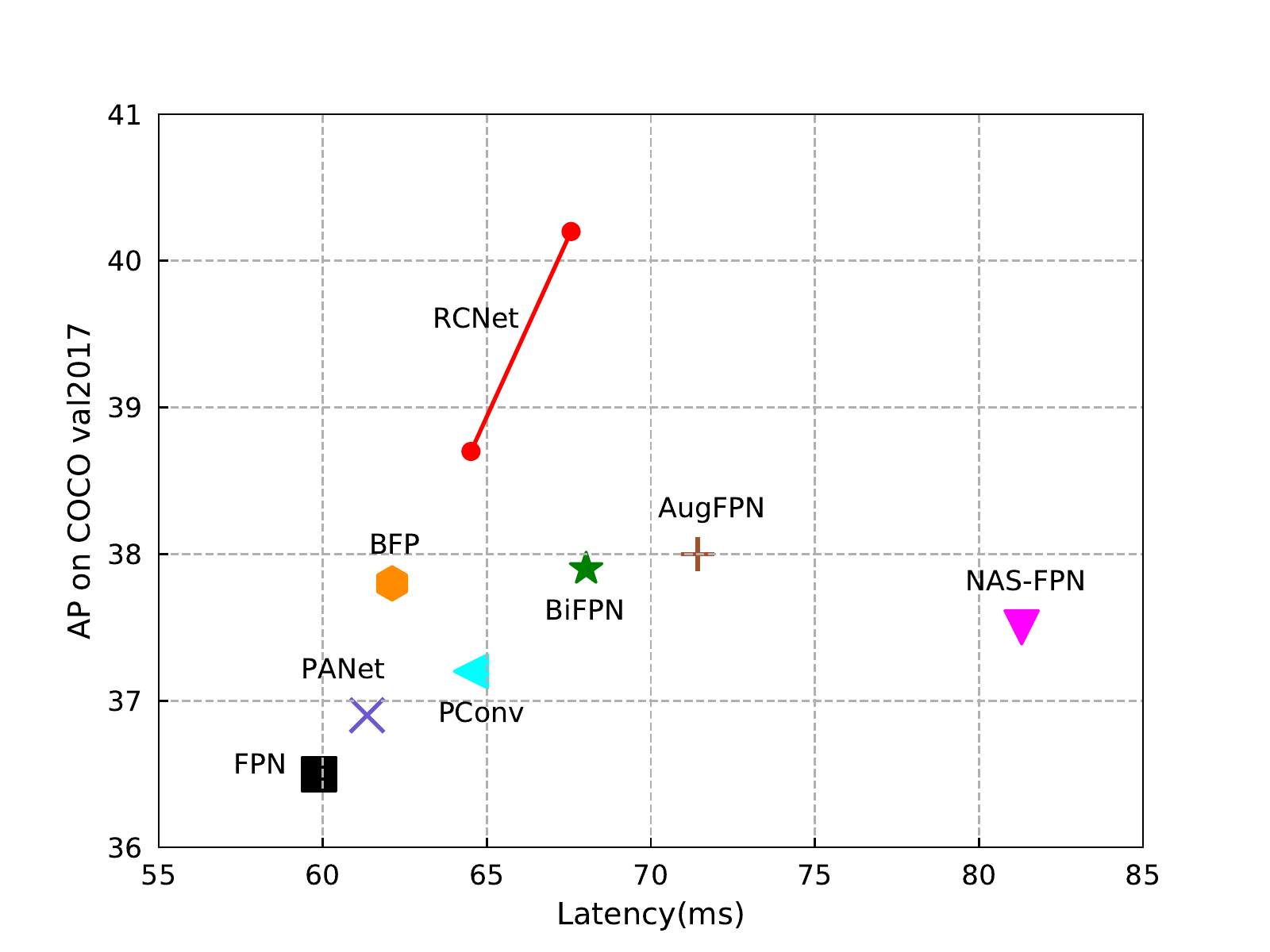}
  \caption{Performance on COCO \textit{val}2017 of various feature fusion methods including FPN \cite{lin2017feature}, PANet \cite{liu2018path}, BFP \cite{pang2019libra}, NAS-FPN \cite{ghiasi2019fpn}, PConv \cite{wang2020scale}, BiFPN \cite{tan2020efficientdet}, AugFPN \cite{guo2020augfpn} and our proposed RCNet. All models are trained for 12 epochs and adopt RetinaNet w/ ResNet-50 as baseline. The left (right) endpoint of red solid line denotes the performance of RCNet without (with) deformable convolution involved.}
  \label{speed-acc}
  \Description{speed-acc}
\end{figure}

\section{Introduction}

\begin{figure*}[t]
  \centering
  \subfigure[FPN]{
    \includegraphics[width=0.22\linewidth]{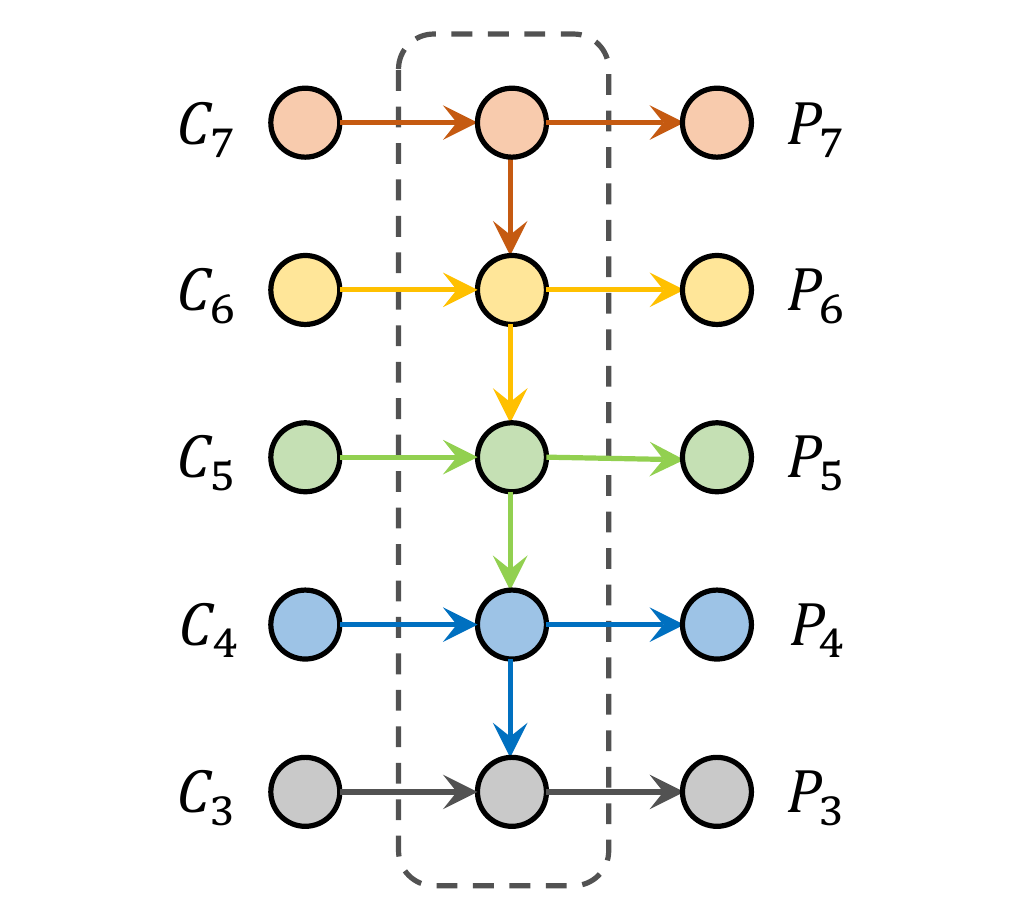}}
    \label{fig:fpna}
  \subfigure[PANet]{
    \includegraphics[width=0.2599\linewidth]{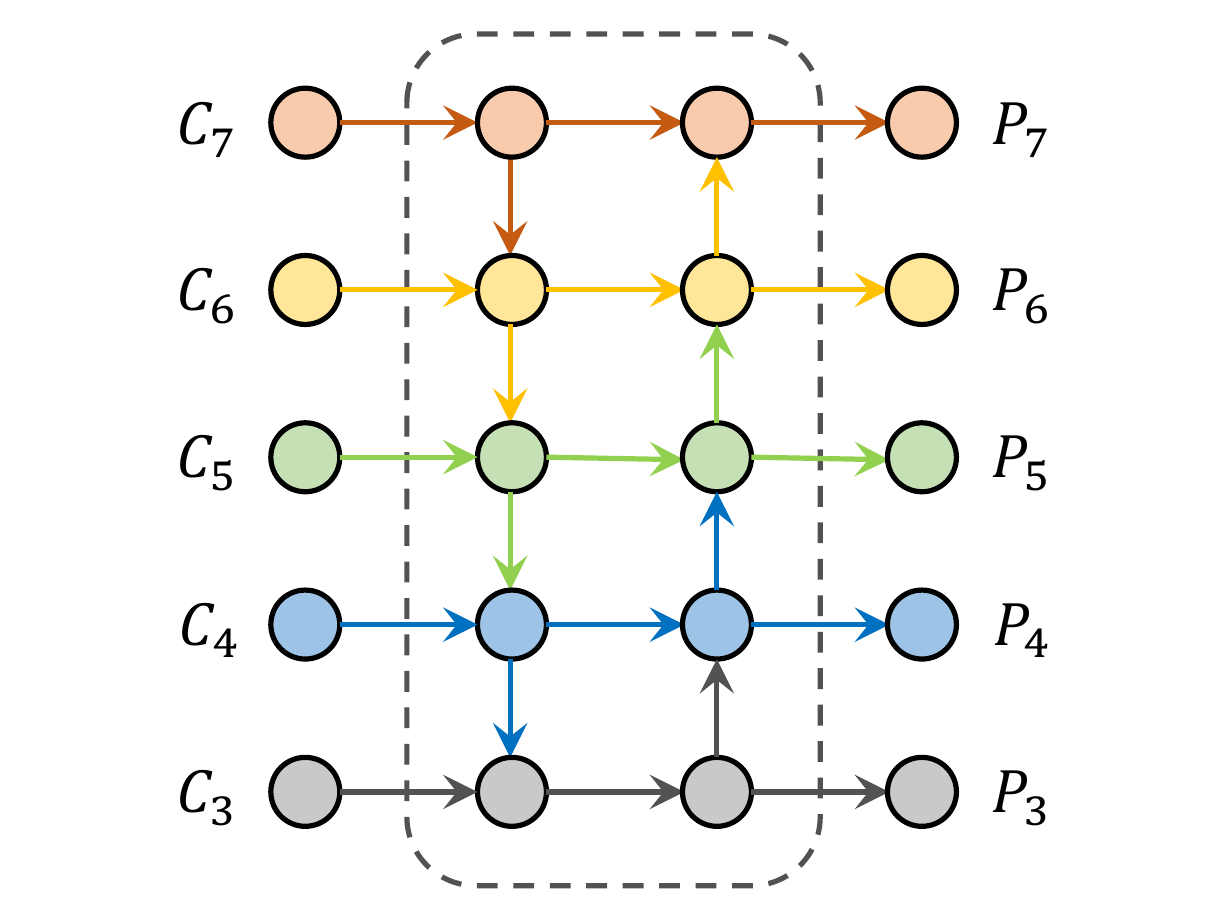}}
    \label{fig:fpnb}
  \subfigure[BiFPN]{
    \includegraphics[width=0.2599\linewidth]{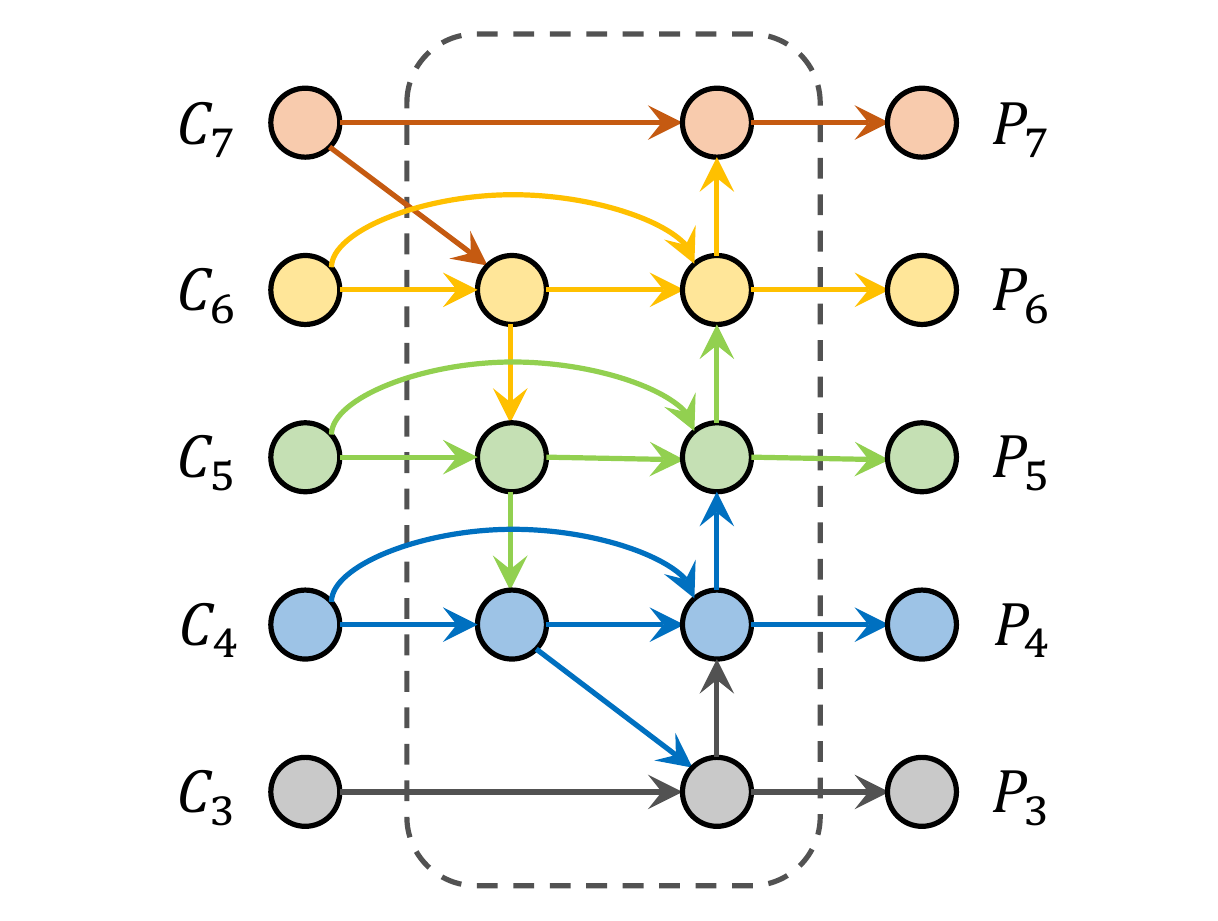}}
    \label{fig:fpnc}
  \subfigure[RevFP (Ours)]{
    \includegraphics[width=0.22\linewidth]{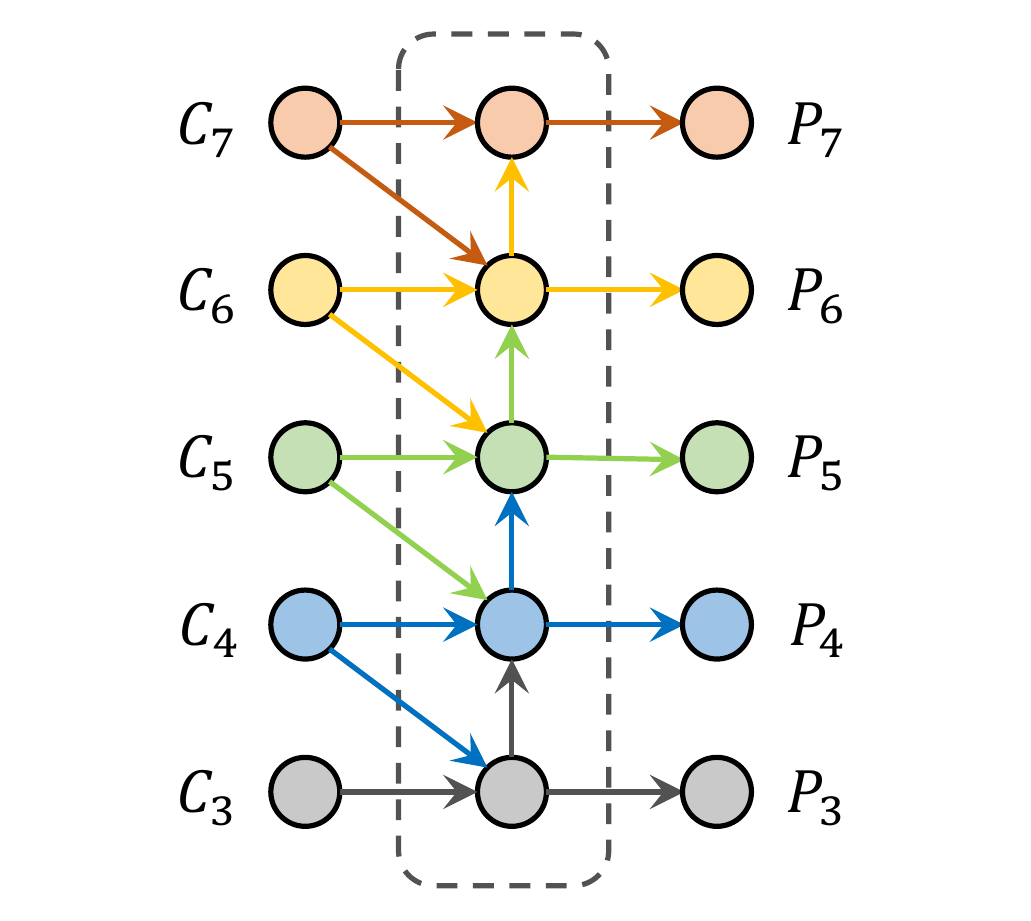}}
    \label{fig:fpnd}
  \caption{Illustration of different feature pyramid designs, including (a) FPN, (b) PANet, (c) BiFPN, (d) RevFP. FPN is limited by the unidirectional top-down information flow. PANet and BiFPN stack multiple feature pyramid layers to allow bidirectional feature fusion. Our RevFP simplifies the bidirectional feature fusion pipeline to improve efficiency and achieve superior performance.}
\end{figure*}

Object detection is one of the most challenging tasks in the field of computer vision. In recent years, with the advance of deep convolutional neural networks, several seminal frameworks for object detection, e.g., \cite{girshick2015fast,ren2016faster,he2017mask,lin2017focal,lin2017feature,redmon2018yolov3} have been proposed where performance grows rapidly. As presented in Figure 2(a), the pioneering work \cite{lin2017feature} builds a top-down pathway with lateral connections to exploit the inherent multi-scale feature representation of deep convolutional network. Inspired by the paradigm of FPN, numerous innovative pyramidal designs for object detection \cite{liu2018path,ghiasi2019fpn,tan2020efficientdet,pang2019libra,guo2020augfpn,zhu2019feature,wang2020scale,qiao2021detectors,kong2018deep,li2020netnet} have been developed. The contributions of these modified pyramidal networks include, but are not limited to: stacking multiple pyramid pathways, dense cross-scale connections, balanced feature representations, scale correlation modeling, and bidirectional fusion. As shown in Figure 2(b), the bidirectional feature pyramid PANet \cite{liu2018path} augments an extra bottom-up pathway to propagate localization information of shallow features to deeper levels to improve large object detection. BiFPN \cite{tan2020efficientdet} simplifies the cross-scale connections in PANet and repeats the same top-down and bottom-up paths multiple times to enable more multi-level feature fusion, as shown in Figure 2(c).

These complicated stacked bidirectional feature pyramids generally yield higher accuracy due to longer feature fusion pipelines. However, the longer pipelines significantly reduce the inference speed of the network. For instance, as shown in Table \ref{tab3}, BiFPN leads to $15\%$ increase of inference time. On the other hand, we find that adding extra local reversed connections mimics the effects of global bidirectional information flow with a negligible decline of inference speed. Based on this observation, we try to simplify the bidirectional pathways paradigm to improve model efficiency. As shown in Figure 2(d), we design a novel single-pathway pyramid, named Reverse Feature Pyramid (RevFP), to bidirectionally integrate features with the minimum efficiency loss and obtain better performance. Different from other bidirectional feature pyramids \cite{liu2018path,tan2020efficientdet}, our proposed RevFP consists of a single bottom-up pathway with augmented local top-down connections. Compared to the top-down pathway in global bidirectional pyramid designs, the local top-down connections only bring insignificant computational burdens. We also tested the pyramid with a top-down pathway and local bottom-up connections but the performance decreased. Moreover, feature-guided upsampling that uses precise low-level localization details to guide upsampling process is adopted by RevFP. We further incorporate the weighted fusion of BiFPN and introduce a simple method to dynamically assign level-wise weights to features of different pyramid levels. 

Another problem of most existing feature pyramids is that: Feature maps of neighboring scales on a feature pyramid correlate the most but the non-neighboring ones hardly correlate \cite{wang2020scale} as local fusion operation focuses more on neighboring scales. To alleviate the discrepancy, Libra-RCNN \cite{pang2019libra} uses the same deeply integrated balanced semantic features to strengthen multi-scale features. The pyramid convolution in SEPC \cite{wang2020scale} conducts 3-D convolution with the kernel size of 3 in the feature pyramid to cater for inter-scale correlation. However, these methods cannot directly interchange information with any other level.

To address this problem, we further introduce Cross-scale Shift Network (CSN), which is a shift-based network and highly complementary to our proposed RevFP. Unlike these aforementioned methods \cite{pang2019libra, wang2020scale}, the multi-scale shift module in CSN directly shifts feature channels to both neighboring and non-neighboring levels to interchange information with any other level. This module is efficient and light-weighted since shift operation is a zero parameter and zero FLOPs operation \cite{wu2018shift}. Besides, we insert a dual global context module to model diverse scale and spatial global context of the shifted features to enhance the correlation among all pyramid levels.

\begin{figure*}[t]
  \centering
  \includegraphics[width=\linewidth]{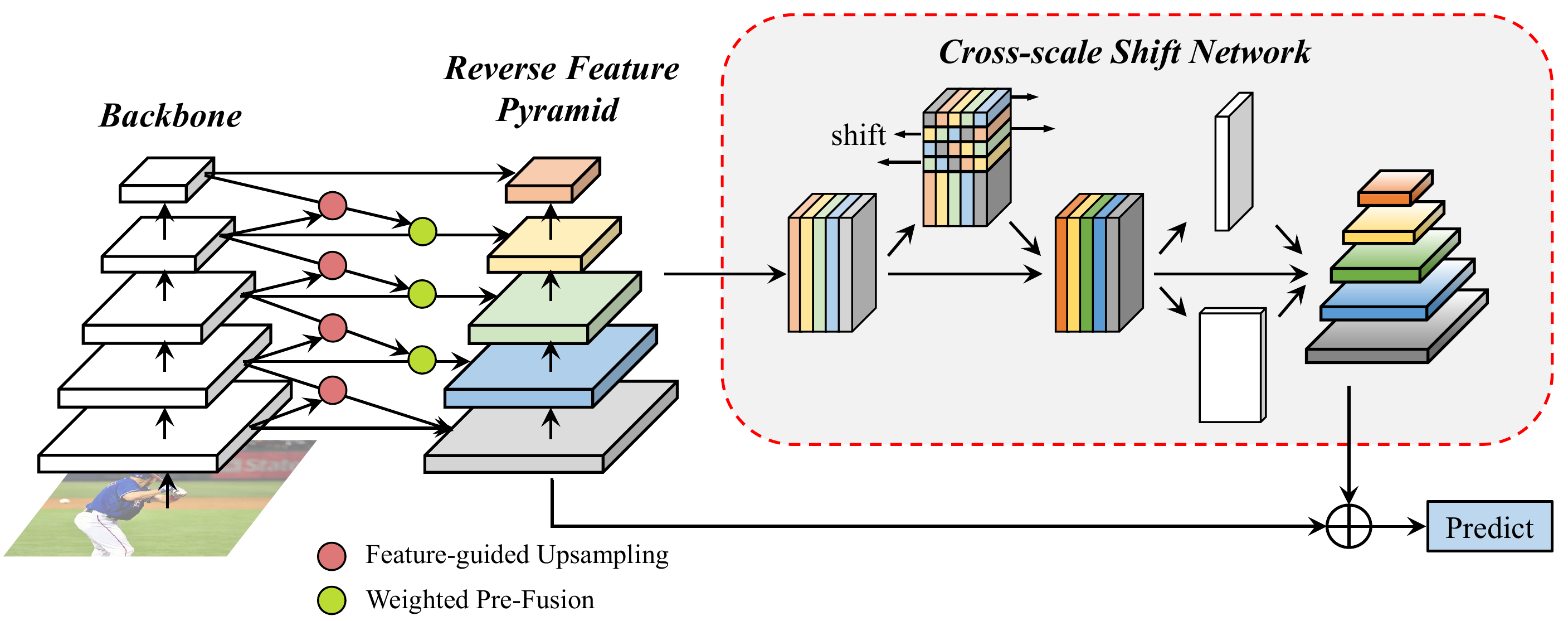}
  \caption{The overall framework of RCNet. $\oplus$ denotes scale-wise addition here. In this paper, our proposed model consists of Reverse Feature Pyramid (RevFP) and Cross-scale Shift Network (CSN). The outputs of these two models are integrated by element-wise addition for subsequent prediction.}
  \Description{Overall framework}
  \label{framework}
\end{figure*}

We combine the proposed RevFP with CSN to form our RCNet, whose overall framework is presented in Figure \ref{framework}. To demonstrate its robustness and effectiveness, we equip RCNet with both single-stage and multi-stage detectors on COCO dataset. RCNet consistently outperforms these competitive baseline methods by 2.8 AP at least. In particular, RCNet with ResNet-50 as backbone achieves as high as 40.2 AP and considerably improves the RetinaNet baseline by 3.7 AP with fewer parameters and negligible latency growth. On COCO \textit{test-dev}, RCNet is able to reach competitive 47.4 AP by using larger backbone ResNet-101 and strong baseline GFL (45.0 AP), even surpassing the state-of-the-art ATSS (45.1 AP) with ResNeXt-101 backbone.

In summary, our contributions in this work are the following:
\begin{itemize}
\item
We propose a local bidirectional feature pyramid network named Reverse Feature Pyramid (RevFP), which achieves better performance and lower latency than other global bidirectional pyramidal designs. 
\item
Cross-scale Shift Network (CSN) is developed to allow more cross-scale features interchange and model both scale and spatial integrated contextual information to further enable all pyramid levels more correlative.
\item
Our proposed RCNet can outperform both one-stage and two-stage baseline methods by a large margin (more than 2.8 AP increase) with negligible additional complexity and achieve very impressive 52.9 AP with the strong backbone (ResNeXt-101-64x4d-DCN) on COCO \textit{test-dev}.
\end{itemize}

\section{Related Works}
\subsection{Object Detection}

Current CNN-based object detection frameworks are almost categorized as one-stage and two-stage detectors. One-stage detectors \cite{lin2017focal,redmon2018yolov3,tian2019fcos,law2018cornernet,duan2019centernet,zhou2019objects} are more efficient but less effective than two-stage ones due to their concise pipelines. They extract features by backbone and build a feature pyramid, directly predict the location and category of objects on pixels of multi-scale features. RetinaNet \cite{lin2017focal} introduces focal loss to address class imbalance during training and pushes the accuracy of one-stage detectors to a new level. FCOS \cite{tian2019fcos} abandons the anchor mechanism and detects objects in a per-pixel prediction fashion. The keypoint-based methods, including CornerNet \cite{law2018cornernet} and CenterNet \cite{zhou2019objects}, model an object as a single keypoint and use keypoint estimation to localize objects. Two-stage object detectors like Fast-RCNN \cite{girshick2015fast}, Faster-RCNN \cite{ren2016faster} and Mask-RCNN \cite{he2017mask} first obtain multiple regions of interest and extract region features. These regions are further fed into the detection head for classification and bounding box regression.

\subsection{Multi-scale Feature Fusion}

Multi-scale feature fusion, which aims to efficiently facilitate cross-scale information integration, plays a crucial role in object detection. The pioneering work \cite{lin2017feature} builds a feature pyramid network, which consists of a top-down pathway and lateral connections to extract multi-level representations. To further enhance the representational capacity of FPN, PANet \cite{liu2018path} augments an extra bottom-up pathway to propagate low-level geometric details. Libra-RCNN \cite{pang2019libra} integrates balanced multi-scale features and refines the original features by adding a non-local block. M2Det \cite{zhao2019m2det} utilizes alternating joint U-Net to extract more representative multi-scale features. NAS-FPN \cite{ghiasi2019fpn} automatically constructs multiple stacked pyramid networks by neural architecture search. Based on PANet, BiFPN \cite{tan2020efficientdet} introduces weighted bidirectional information flow. Different from the aforementioned methods, SEPC \cite{wang2020scale} conducts pyramid convolution inside the feature pyramid to capture cross-scale correlation. 

\subsection{Shift-based Network}

As a parameter-free and FLOP-free operation, the shift operation is introduced to improve the performance of CNNs while reducing model complexity in previous work. The shift operation in ShiftNet \cite{wu2018shift} is designed as an efficient alternative to expensive spatial convolutions to aggregate spatial information. FE-Net \cite{chen2019all} develops a novel component called Sparse Shift Layer (SSL) to suppress redundant shift operations by adding displacement penalty during optimization. Three shift-based primitives presented in AddressNet \cite{he2019addressnet} significantly reduce the inference time by minimizing the memory copy of shift operations. Meanwhile, the work \cite{lin2019tsm} proposes Temporal Shift Module (TSM) to efficiently model both temporal and spatial information. TSM can exchange features among neighboring frames in 2D CNNs by shifting channels along the temporal dimension. Recently, inspired by hardware implementation of multiplication, You et al. \cite{you2020shiftaddnet} propose ShiftAddNet, which consists of only bit-shift and additive weight layers. 

\section{Reverse Feature Pyramid}

\subsection{Pyramid Pathway Design}

In this section, we first revisit the top-down pathway design of FPN. FPN propagates deep high-level features with more semantic meanings to the adjacent lower level via a top-down pathway. Let $C_i$ and $P_i$ (the resolution is $1/2^i$ of the input images) denote the feature map of $i$-th stage of the backbone network and output of the feature pyramid at level $i$, respectively. $f_i$ is the $i$-th multi-scale feature fusion operation. The output $P_i$ in FPN is calculated as follows:
\begin{equation}
  P_i=f_i(C_i,P_{i+1}),   
\end{equation}
where $i \in \{l_{min},\cdots,l_{max}\}$. In our experiments, we set $l_{min}=3,l_{max}=7$ for one-stage detectors and $l_{min}=2,l_{max}=6$ for two-stage detectors. The top-down FPN is inherently limited by the one-way information flow as the fusion process at level $i$ only requires feature maps generated by deeper layers of backbone, ignoring features at the lower level. 

Recent works \cite{liu2018path,tan2020efficientdet} augmented an extra bottom-up pathway to significantly enhance accurate information flow. The fusion operation in these stacked top-down and bottom-up pathways is shown as $P_{i,0} = C_i$ and
\begin{equation}
    P_{i,j}=\left\{
    \begin{aligned}
    f_{i,j}(P_{i,j-1},P_{i+1,j}) &,& j=1,3,\cdots, \\
    f_{i,j}(P_{i,j-1},P_{i-1,j}) &,& j=2,4,\cdots, \\
    \end{aligned}
    \right.
\end{equation}
where $j$ denotes the index of feature pyramid layer. It is proved multiple ordered top-down pathways and bottom-up pathways essentially enhance the representational capacity of feature pyramid but lead to longer pipelines that bring new challenges: parameters growth and inference speed decline. 

To efficiently enable bidirectional feature fusion, we propose a single-pathway pyramid named Reverse Feature Pyramid (RevFP). Different from other feature pyramid methods, the local fusion operation of RevFP simultaneously integrates both high-level and low-level representations. The fusion operation in the bottom-up pathway of RevFP is described as:
\begin{equation}
  P_i=f_{i}(C_i,C_{i+1},P_{i-1}),
\end{equation}
where high-level feature $C_{i+1}$ is propagated through the local top-down connection. The local top-down connections are very critical to the bottom-up design. 

\subsection{Feature-guided Upsampling}

Following the conventional FPN, we upsample the spatial resolution of a coarser-resolution feature map by a factor of 2 before performing fusion. Low-level features are usually deemed rich in containing geometric patterns, which is contrary to high-level features \cite{zeiler2014visualizing} Inspired by this, we introduce a very light-weighted and efficient upsampling operation named feature-guided upsampling. This operation uses the accurate localization details in shallow low-level features to guide the upsampling process to lead the coarser features more discriminative before weighted pre-fusion. To be specific, the spatial resolution of $C_{i+1}$ is upsampled by bilinear interpolation at first. We further use concatenation operation followed by a $3 \times 3$ convolution to reduce the feature channel to 1 and obtain the spatial weights, which are then normalized by softmax function. Similar to \cite{vaswani2017attention}, scaling operation with scaling factor of $\frac{T}{\sqrt{d_{i}}}$ is added before the softmax function to make the training process stable. In our experiments, ${d_{i}}$ is set as 256 for all pyramid levels. We adopt the learnable temperature parameter $T$ (initialized as 1) since the receptive fields of features at different scales vary enormously. Turning learnable $T$ to a constant slightly degrades the final performance. We rescale the spatial attention maps after normalization with a scaling factor of $h_{i} * w_{i}$ to ensure the average spatial weight is 1. Finally, the spatial weights are multiplied to all the activations along the spatial axis of $C_{i+1}$ for the corresponding positions. 

\subsection{Dynamic Weighted Fusion}

We incorporate the weighted fusion method introduced in \cite{tan2020efficientdet} based on the bottom-up design and propose dynamic weighted fusion strategy to fully integrate both low-level patterns and high-level semantics. The dynamic weighted fusion process of RevFP is divided into weighted pre-fusion and weighted post-fusion in our implementation. During weighted pre-fusion, we calculate the weights of input features $C_{i}$ and $C_{i+1}$, where $i \in \{l_{min},\cdots,l_{max}\}$ denotes pyramid level, $C_i$ is the input feature map at level $i$ and $C_{i+1}$ is the high-level feature map recalibrated by feature-guided upsampling. Specifically, we apply concatenation operation to both $C_i$ and $C_{i+1}$ to capture the cross-scale correlation \cite{wang2018non} and apply global average pooling layer to notably decrease the computation cost. Then $1 \times 1$ convolution followed by sigmoid function is used to generate the dynamic level-wise weight $\mathcal W'_{i}\in \mathbb{R}^{1 \times 1 \times 1}$ of $C_i$ and the weight of $C_{i+1}$ is set as $1-\mathcal W'_{i}$ for simplification. The intermediate feature $P'_i$ after the first step fusion is calculated by the following formula:
\begin{equation}
  P'_i=Conv(\mathcal{W}'_{i}*C_i+(1-\mathcal{W}'_{i})*C_{i+1}),
\end{equation}
where $Conv$ denotes $3 \times 3$ convolution. As for the weighted post-fusion that follows weighted pre-fusion, $P'_i$ and $P_{i-1}$ (downsampled by max pooling to reduce parameters) are the input features and the final output $P_i$ is constructed in a similar manner:
\begin{equation}
  P_i=Conv(\mathcal{W}_{i}*P'_i+(1-\mathcal{W}_{i})*P_{i-1}).
\end{equation}
We directly calculate the output feature $P_i$ by $P_i=P'_i$ at the bottom-most level $l_{min}$. Equation 4 can be rewritten as $P'_i=C_i$ for the top-most pyramid level $l_{max}$. This two-step fusion strategy is not as necessary as the structure design to RevFP compared to one-step weighted fusion. In our implementation, we use deformable convolution followed by batch normalization as $Conv$ to further model geometric transformations of different scales.

\section{Cross-scale Shift Network}

The output $P_{i}$ of RevFP only directly merges features at level $i-1$, $i$ and $i+1$, the information from other levels (e.g., $l_{min}$) is diluted due to the local fusion operation. The motivation we design Cross-scale Shift Network (CSN) is simple: Scale shift operation can serve as a parameter-free and FLOP-free global fusion operation that is complementary to the local fusion in RevFP to facilitate multi-scale fusion without much computational burden. Besides, integrated context information from each level captured by CSN further enables multi-scale features more correlative.

\subsection{Pipeline}

The feature pyramid $\{P_{i}\mid i=l_{min},\cdots,l_{max} \}$ constructed by RevFP is further fed into CSN. CSN is composed of two modules: multi-scale shift module and dual global context module. $\{P_{i}\mid i=l_{min},\cdots,l_{max} \}$ is resized by interpolation or pooling to the same resolutions $h_{k}*w_{k}$ (e.g., $k=4$) and merged as a sequence of features $P \in \mathbb{R}^{d_{k} \times n \times h_{k} \times w_{k}}$, where $n=1+l_{max}-l_{min}$. CSN receives $P$ as input and directly fuses shallow and deep features across levels. The output $\hat{P} \in \mathbb{R}^{d_{k} \times n \times h_{k} \times w_{k}}$ is later divided into features of $n$ levels and resized to the corresponding resolutions ($h_{i}*w_{i}$) to construct another feature pyramid $\{\hat{P}_{i}\mid i=l_{min},\cdots,l_{max} \}$. Finally, we combine the outputs of RevFP and CSN to get a fused feature pyramid $\{P_{i}+\hat{P}_{i}\mid i=l_{min},\cdots,l_{max} \}$ with stronger representations for subsequent classification and regression.

\subsection{Multi-scale Shift Module}

\begin{figure}[ht]
  \centering
  \includegraphics[width=\linewidth]{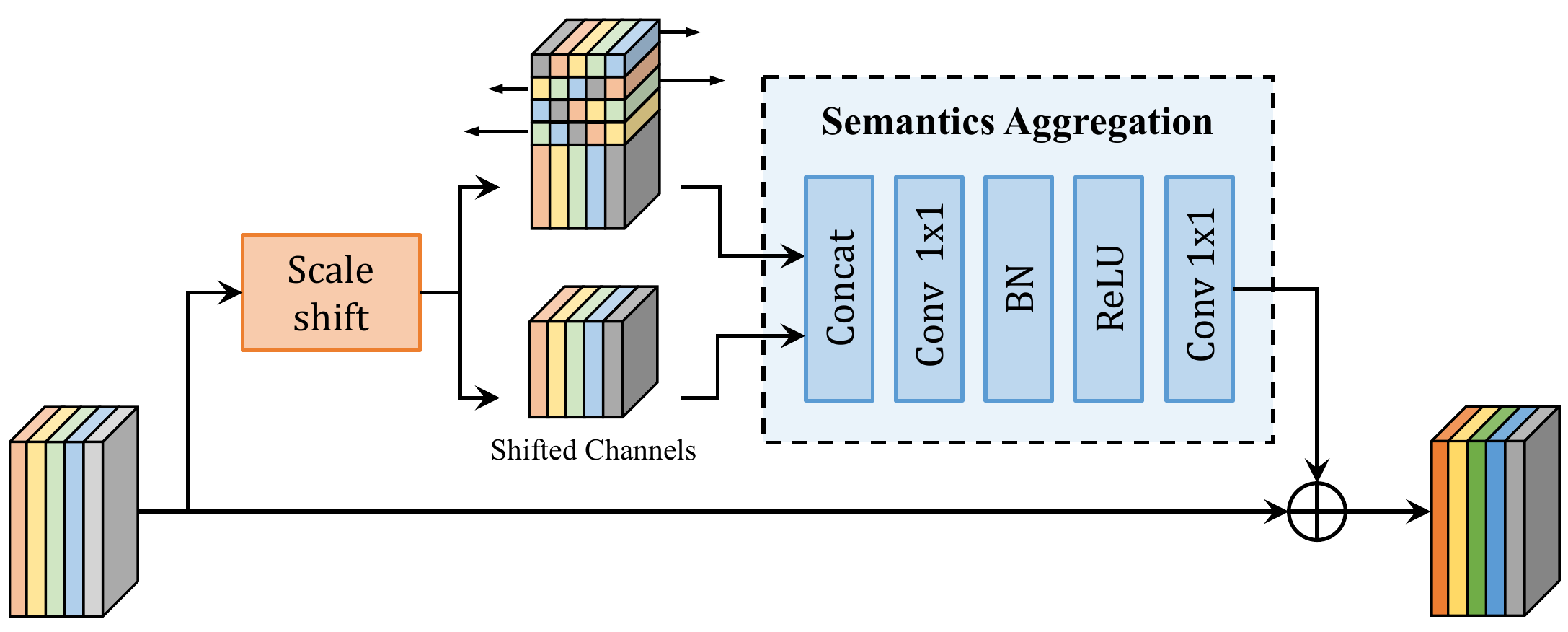}
  \caption{Architecture of our proposed multi-scale shift module. $\oplus$ denotes element-wise addition. We set the numbers of pyramid levels of input as 5 in the figure. Different colors of features denote different scales.}
  \Description{CSN Overall framework}
  \label{msm}
\end{figure}

Unlike the previous shift-based networks \cite{wu2018shift,chen2019all,lin2019tsm,you2020shiftaddnet}, as shown in Figure \ref{msm}, our proposed multi-scale shift module allows more multi-scale feature fusion by shifting subsets of channels along the scale dimension. The features of different scales are exchanged among both adjacent and non-adjacent levels.

The input feature pyramid $P$ can be described as the ordered sequence $\{P_{i}\mid i=l_{min},\cdots,l_{max} \}$. We first calculate $P^{-2}$, $P^{-1}$, $P^0$, $P^{+1}$ and $P^{+2}$ by circularly shifting $P$ to right and left by 0, 1, 2 along scale dimension. And the ordered weight sequence $W=\{W_{i}\mid i=1,\cdots,5 \}$ is multiplied by the shifted sequences and sum up to be $Y$, which is defined by:
\begin{equation}
  Y = \sum_{j=1}^5W_{j}*P^{j-3}.
\end{equation}
$Y$ can be split into $\{Y_{i}\mid i=l_{min},\cdots,l_{max} \}$, $Y_{i}$ is calculated by:
\begin{equation}
  Y_{i} = \sum_{j=1}^5W_{j}*P_{i+j-3}.
\end{equation}
Note that we follow the circulant padding strategy to pad the empty boundary. As a concrete example, we set $l_{min}=3, l_{max}=7$ and obtain $Y_{6}$ by:
\begin{equation}
  Y_{6} = W_{1}*P_{4}+W_{2}*P_{5}+W_{3}*P_{6}+W_{4}*P_{7}+W_{5}*P_{3}.
\end{equation}
This process also can be viewed as a convolution with the kernel size of 5 and circulant padding of 2 along scale dimension. However, shift operation enables information interchange at zero computation cost. Following previous work \cite{lin2019tsm}, we only shift part of feature channels and adopt residual connection to reduce data movement. Furthermore, the shifted channels are merged into current features by concatenation to restore the information contained in shifted channels for the current level. Experimentally, we find our model achieves the best performance when shift ratio $r$ is 4 (1/4 channels are shifted). Separate $1 \times 1$ convolution layers are added for channel reduction and multi-level semantics aggregation, serving as aforementioned weight $W$. Batch normalization and ReLU activation function are inserted between these two convolution layers to ease optimization. 

\subsection{Dual Global Context}

To make the shifted features at different levels more correlative, we exploit the integrated contextual information by inserting a dual global context module. This module consists of two branches: scale context branch and spatial context branch. In the scale context branch, we first squeeze global spatial information at multiple pyramid levels by global average pooling. A $1 \times 1$ convolution is used to obtain channel context of different scales, which is further merged into multi-scale features by softmax scaling. Then we perform average pooling along scale dimension and apply linear transformation ($1 \times 1$ convolution) to capture the global scale context. The spatial context branch is similar to the scale context branch except for the sequence of operations which is converted from ``spatial pooling, channel context fusion, scale pooling'' to ``scale pooling, channel context fusion, spatial pooling''. The outputs of these two branches are finally added back to the main stream.

\section{Experiments}

\subsection{Dataset and Evaluation Metrics}

Our experiments are conducted on the challenging MS COCO 2017 dataset. The training set (\textit{train}2017) that consists of 115K labeled images is utilized for training. We report the detection results by default on the validation set (\textit{val}2017) which has 5K images. To compare with the state-of-art approaches, results evaluated on the test set (\textit{test-dev}) are also reported. All results follow the standard COCO Average Precision(AP) metrics.

\subsection{Implementation Details}

\begin{table*}[t]
  \caption{Results of different single-stage and multi-stage architectures on COCO \textit{val}2017. "BBox" denotes the object detection task and "Segm" stands for the instance segmentation task. For a fair comparison, we report the results of corresponding baseline methods implemented by mmdetection.
}
  \label{tab2}
  \begin{tabular}{cccccccccc}
    \toprule
    Detector& Note&  Task& AP& AP$_{50}$& AP$_{75}$& AP$_{S}$& AP$_{M}$& AP$_{L}$\\
    \midrule
    RetinaNet& baseline& BBox& 36.5& 55.4& 39.1& 20.4& 40.3& 48.1 \\
    RetinaNet& RCNet& BBox& \textbf{40.2}& \textbf{59.6}& \textbf{43.1
    }& \textbf{23.7}& \textbf{44.2}& \textbf{53.8} \\
    ATSS& baseline& BBox& 39.4& 57.6& 42.8& 23.6& 42.9& 50.3 \\
    ATSS& RCNet& BBox& \textbf{42.6}& \textbf{60.3}& \textbf{46.3}& \textbf{25.6}& \textbf{46.3}& \textbf{55.9} \\
    GFL& baseline& BBox& 40.2& 58.4& 43.3& 23.3& 44.0& 52.2 \\
    GFL& RCNet& BBox& \textbf{43.1}& \textbf{60.9}& \textbf{47.0}& \textbf{25.1}& \textbf{47.2}& \textbf{57.5} \\
    \hline
    Faster R-CNN& baseline& BBox& 37.4& 58.1& 40.4& 21.2& 41.0& 48.1 \\
    Faster R-CNN& RCNet& BBox& \textbf{40.2}& \textbf{60.9}& \textbf{43.6}& \textbf{25.0}& \textbf{43.5}& \textbf{52.9} \\
    Mask R-CNN& baseline& BBox& 38.2& 58.8& 41.4& 21.9& 40.9& 49.5 \\
    Mask R-CNN& RCNet& BBox& \textbf{40.7}& \textbf{60.9}& \textbf{44.5}& \textbf{24.7}& \textbf{44.1}& \textbf{53.0} \\
    Mask R-CNN& baseline& Segm& 34.7& 55.7& 37.2& 18.3& 37.4& 47.2 \\
    Mask R-CNN& RCNet& Segm& \textbf{36.2}& \textbf{57.7}& \textbf{38.6}& \textbf{20.5}& \textbf{39.0}& \textbf{49.7} \\
    \bottomrule
  \end{tabular}
\end{table*}

All experiments are implemented on mmdetection \cite{chen2019mmdetection} and all hyper-parameters follow the default settings in mmdetection for fair comparisons. More implementation details can be found in the supplementary material.

\subsection{Main Results}

In this subsection, we show the main results on COCO \textit{val}2017 in Table \ref{tab2}. We use ResNet-50 as the backbone and apply RCNet to both one-stage and two-stage object detectors to investigate its effectiveness and robustness. The one-stage detectors we adopted includes RetinaNet \cite{lin2017focal} and powerful baselines: ATSS \cite{zhang2020bridging} and GFL \cite{li2020generalized}. The upper part of the results presents one-stage object detectors. Compared to FPN, RCNet boosts the accuracy of anchor-based model RetinaNet by 3.7 AP, which is a dramatic improvement. Moreover, RCNet consistently outperforms the baselines by a large margin on the state-of-the-art one-stage detectors, achieving 42.6 and 43.1 AP on ATSS and GFL, respectively. It is worth noting that RCNet invariably obtains more than 5 AP$_{L}$ gain on different one-stage detectors. This validates that the bottom-up information flow vastly reinforces the model's detection performance, especially on large objects.

As for two-stage detection frameworks, we further incorporate our RCNet into Faster R-CNN \cite{ren2016faster} and Mask R-CNN \cite{he2017mask}. The results are illustrated in the lower part of Table \ref{tab2}. Obviously, Faster R-CNN achieves 40.2 AP and notably surpasses the performance of baseline by replacing FPN with RevFP and CSN. The improvements of RCNet on Mask R-CNN are also remarkable, reaching 40.7 bbox AP on object detection. In regards to instance segmentation, Mask R-CNN is improved by 1.5 mask AP. These results of different single-stage and multi-stage detectors prove RCNet can consistently bring great improvements on various detectors and tasks.

\subsection{Comparison with Other Feature Fusion Methods}

In this subsection, we compare RCNet with other state-of-the-art feature fusion methods, including FPN \cite{lin2017feature}, PANet \cite{liu2018path}, PConv \cite{wang2020scale}, AugFPN \cite{guo2020augfpn}, BFP \cite{pang2019libra}, NAS-FPN \cite{ghiasi2019fpn} and BiFPN \cite{tan2020efficientdet}. Table \ref{tab3} presents the model size, inference latency, and accuracy of these models. We stack 7 merging-cells in NAS-FPN and set the BiFPN depth as 4. Results of "RCNet w/o DCN" where vanilla $3 \times 3$ convolution layers are utilized for fusion are also reported. It is surprising that ``RCNet w/o DCN'' boosts the baseline by absolute 2.2 AP, which is even superior to other state-of-the-art feature fusion methods. As for the efficiency, the parameters of ``RCNet w/o DCN'' are smaller than FPN and the latency only increases by around $7\%$. Simultaneously, RCNet dramatically benefits RetinaNet by 3.7 AP and greatly surpasses other state-of-the-art feature fusion methods with deformable convolutions involved. As shown in Figure \ref{speed-acc}, RCNet achieves better speed-accuracy trade-off than other competitive counterparts.

\begin{table}[ht]
  \caption{Performance of different feature fusion methods on RetinaNet with ResNet-50 backbone. Results are evaluated on COCO \textit{val}2017.
  }
  \label{tab3}
  \begin{tabular}{cccccc}
    \toprule
    Method& Params& Latency& AP& AP$_{50}$& AP$_{75}$\\
    \midrule
    FPN & 37.7M& 59.9ms& 36.5& 55.4& 39.1\\
    PANet & 40.1M& 61.4ms& 36.9& 56.0& 39.5\\
    PConv & 41.3M& 64.5ms& 37.2& 57.0& 39.5\\
    NAS-FPN & \textbf{59.7M}& \textbf{81.3ms}& 37.5& 55.4& 40.0\\
    BFP & 38.0M& 62.1ms& 37.8& 56.9& 40.1\\
    BiFPN & 34.2M& 68.0ms& 37.9& 56.8& 40.7\\
    AugFPN & 53.8M& 70.9ms& 38.0& 57.7& 40.5\\
    \hline
    RCNet w/o DCN & 36.4M& 64.5ms& 38.7& 58.3& 41.2\\
    RCNet & 37.4M& 67.6ms& \textbf{40.2}& \textbf{59.6}& \textbf{43.1}\\
  \bottomrule
\end{tabular}
\end{table}

\subsection{Comparison with State-of-the-art Detectors}

\begin{table*}[t]
  \caption{Comparing RCNet against state-of-the-art methods on COCO \textit{test-dev}. Results are arranged in the increasing order of AP. The symbol $^\dagger$ and $^\ddagger$ indicate the multi-scale training and the multi-scale testing, respectively.
}
  \label{tab4}
  \begin{tabular}{l|l|c|ccc|ccc|c}
    \toprule
    Method& Backbone&  Epochs& AP& AP$_{50}$& AP$_{75}$& AP$_{S}$& AP$_{M}$& AP$_{L}$& Reference\\
    \midrule
    \textit{single-stage:}& & & & & & & & & \\
    CornerNet$^\dagger$ \cite{law2018cornernet}& Hourglass-104& 200& 40.6& 56.4& 43.2& 19.1& 42.8& 54.3& ECCV18 \\
    FoveaBox$^\dagger$ \cite{kong2020foveabox}& ResNeXt-101& 18& 42.1& 61.9& 45.2& 24.9& 46.8& 55.6& TIP20 \\
    FSAF$^\dagger$ \cite{zhu2019feature}& ResNeXt-101-64x4d& 24& 42.9& 63.8& 46.3& 26.6& 46.2& 52.7& CVPR19 \\
    FCOS$^\dagger$ \cite{tian2019fcos}& ResNeXt-101-64x4d& 24& 44.7& 64.1& 48.4& 27.6& 47.5& 55.6& ICCV19 \\
    CenterNet$^\dagger$ \cite{duan2019centernet}& Hourglass-104& 190& 44.9& 62.4& 48.1& 25.6& 47.4& 57.4& ICCV19 \\
    FreeAnchor$^\dagger$ \cite{zhang2021learning}& ResNeXt-101-32x8d& 24& 44.9& 64.3& 48.5& 26.8& 48.3& 55.9& NIPS19 \\
    ATSS$^\dagger$ \cite{zhang2020bridging}& ResNeXt-101-32x8d& 24& 45.1& 63.9& 49.1& 27.9& 48.2& 54.6& CVPR20 \\
    CentripetalNet$^\dagger$ \cite{dong2020centripetalnet}& Hourglass-104& 210& 45.8& 63.0& 49.3& 25.0& 48.2& 58.7& CVPR20 \\
    SAPD$^\dagger$ \cite{zhu2019soft}& ResNeXt-101-64x4d-DCN& 24& 47.4& 67.4& 51.1& 28.1& 50.3& 61.5& ECCV20 \\
    SEPC$^\dagger$ \cite{wang2020scale}& ResNeXt-101-64x4d& 24& 47.7& 67.3& 51.7& 29.2& 50.8& 60.3& CVPR20 \\
    GFL$^\dagger$ \cite{li2020generalized}& ResNeXt-101-32x4d-DCN& 24& 48.2& 67.4& 52.6& 29.2& 51.7& 60.2& NIPS20 \\
    NAS-FPN \cite{ghiasi2019fpn}& AmoebaNet& 150& 48.3& -& -& -& -& -& CVPR19 \\
    PAA$^\dagger$ \cite{kim2020probabilistic}& ResNeXt-101-64x4d-DCN& 24& 49.0& 67.8& 53.3& 30.2& 52.8& 62.2& ECCV20 \\
    \hline
    \textit{multi-stage:}& & & & & & & & & \\
    Grid R-CNN \cite{lu2019grid}& ResNet-101& 20& 41.5& 60.9& 44.5& 23.3& 44.9& 53.1& CVPR19 \\
    Cascade R-CNN \cite{cai2018cascade}& ResNet-101& 18& 42.8& 62.1& 46.3& 23.7& 45.5& 55.2& CVPR18 \\
    Libra R-CNN \cite{pang2019libra}& ResNeXt-101-64x4d& 12& 43.0& 64.0& 47.0& 25.3& 45.6& 54.6& CVPR19 \\
    TSD \cite{song2020revisiting}& ResNet-101& 20& 43.2& 64.0& 46.9& 24.0& 46.3& 55.8& CVPR20 \\
    RepPoints$^\dagger$ \cite{yang2019reppoints}&  ResNet-101-DCN& 24& 45.0& 66.1& 49.0& 26.6& 48.6& 57.5& ICCV19 \\
    TridentNet$^\dagger$ \cite{li2019scale}& ResNet-101-DCN& 36& 46.8& 67.6& 51.5& 28.0& 51.2& 60.5& ICCV19 \\
    HTC \cite{chen2019hybrid}& ResNeXt-101-64x4d& 20& 47.1& 63.9& 44.7& 22.8& 43.9& 54.6& CVPR19 \\
    RepPoints v2$^\dagger$ \cite{chen2020reppoints}& ResNeXt-101-64x4d& 24& 47.8& 67.3& 51.7& 29.3& 50.7& 59.5& NIPS20 \\
    BorderDet$^\dagger$ \cite{qiu2020borderdet}& ResNeXt-101-64x4d-DCN& 24& 48.0& 67.1& 52.1& 29.4& 50.7& 60.5& ECCV20 \\
    \hline
    RCNet$^\dagger$& ResNet-101& 24& 47.4& 65.8& 51.4& 28.5& 51.3& 59.3& MM21 \\
    RCNet$^\dagger$& ResNeXt-101-64x4d& 24& 49.2& 67.8& 53.6& 30.4& 52.9& 61.5& MM21 \\
    RCNet$^\dagger$& ResNeXt-101-64x4d-DCN& 24& 50.5& 69.1& 55.0& 30.9& 53.9& 63.9& MM21 \\
    RCNet$^{\dagger}$$^{\ddagger}$& ResNeXt-101-64x4d-DCN& 24& \textbf{52.9}& \textbf{70.5}& \textbf{58.4}& \textbf{35.4}& \textbf{55.7}& \textbf{65.5}& MM21 \\
    \bottomrule
  \end{tabular}
\end{table*}

We report the results of RCNet on COCO \textit{test-dev} and compare it with other state-of-the-art object detection approaches. We replace FPN in the competitive baseline GFL with RevFP and CSN, use stronger backbones such as ResNet-101 and ResNeXt-101. As the batch size per GPU is 1, we replace all BN layers in the pyramid with GN \cite{wu2018group}, whose performance is inferior to BN when using ResNet-50. The multi-scale training strategy is adopted and the short edge of an image is randomly sampled from [480, 960] during training.  We train models for 24 epochs and multiply the learning rate by 0.1 after 16 and 22 epochs. Results of all models obtained with a single model and without test-time augmentation are reported. We further show the results of our best model with multi-scale testing. As shown in Table \ref{tab4}, We find that RCNet achieves an AP of 47.4 when using ResNet-101 as backbone, even surpasses some advanced detectors (e.g., ATSS \cite{zhang2020bridging}) with ResNeXt-101 backbone. By using more powerful ResNeXt-101 as the backbone, the AP is boosted to 49.2 without bells and whistles. RCNet with deformable backbone \cite{dai2017deformable} reaches 50.5 AP, which is superior to other state-of-the-art detectors. Finally, our best model achieves impressive 52.9 AP with the multi-scale testing strategy.

\subsection{Ablation Study}
\noindent\textbf{Effect of each component.}
We first perform a comprehensive component-wise ablation study on RetinaNet, where different components are omitted to thoroughly analyze the effect of each component. As shown in Table \ref{tab5}, by replacing FPN with RevFP, RCNet achieves 40.2 AP, which leads to an absolute 2.7 AP gain. RCNet still makes significant improvements (2.2 AP increase) with vanilla $3 \times 3$ convolution instead of deformable convolution. We also find the feature-guided upsampling and the dynamic weighted fusion both benefit the performance by 0.5 AP and bring negligible extra parameters and computation costs by presenting the performance of ``RCNet w/o guide'' and ``RCNet w/o weight''. 

For CSN, we get a drop of 1.4 AP by omitting it as in ``RCNet w/o CSN''. Specifically, the decrease of AP$_{S}$, which is 3.3 points lower than RCNet, contributes most to the final performance deterioration. This demonstrates that CSN is able to considerably boost the performance of small object detection as well as improving the AP of medium objects and large objects by 1.0 points. Then, we show the insertion of the multi-scale shift module in CSN brings 0.6 AP improvements in ``RCNet w/o shift''. Finally, results in ``RCNet w/o context'' prove that our proposed light-weighted dual global context, benefits the model by 0.8 AP. The results in Table \ref{tab5} manifest that all components in RCNet are complementary and indispensable.
\begin{figure*}[t]
  \centering
  \includegraphics[width=\linewidth]{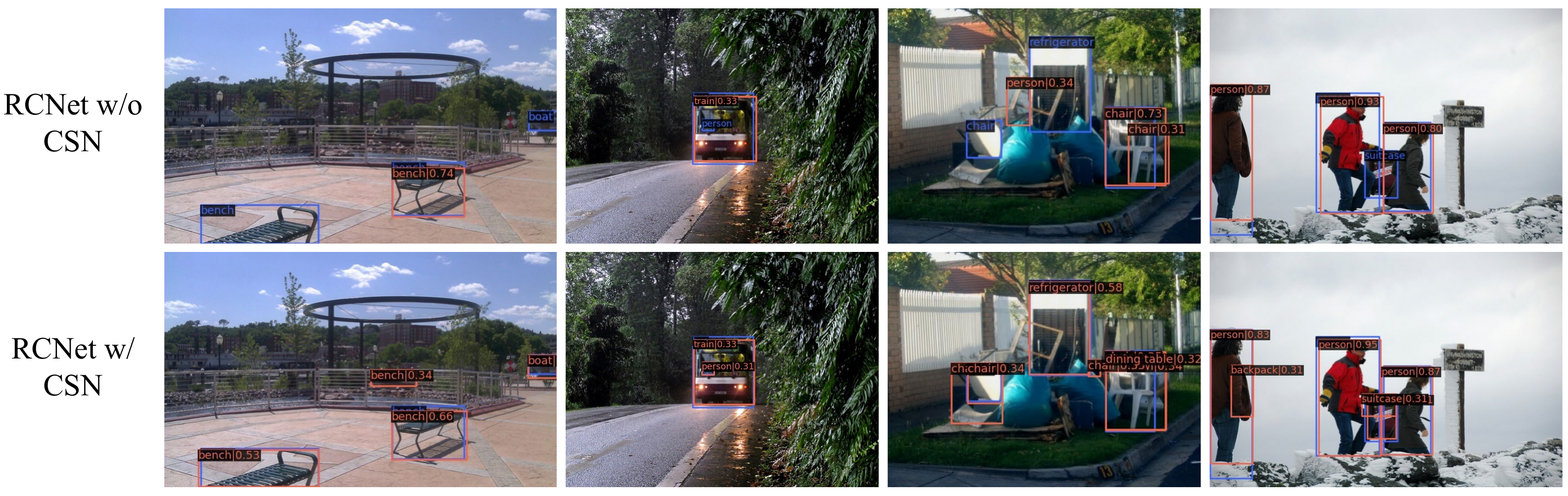}
  \caption{Visualization of detection results of RCNet w/o CSN and RCNet w/ CSN. Both models are built upon ResNet-50 on COCO \textit{val}2017. Blue bounding boxes denote ground truth and orange boxes denote the predictions.}
  \Description{CSNVis}
  \label{csnvis}
\end{figure*}
\begin{table}[ht]
  \caption{Detection performance after individually omitting each component of RCNet. Results evaluated on COCO \textit{val}2017 are reported.}
  \label{tab5}
  \begin{tabular}{l|cccccc}
    \toprule
    Method& AP& AP$_{S}$& AP$_{M}$& AP$_{L}$ \\
    \midrule
    baseline & 36.5& 20.4& 40.3& 48.1 \\
    RCNet & 40.2& 23.7& 44.2& 53.8 \\
    \hline
    RCNet w/o RevFP & 37.5& 22.5& 41.6& 47.8 \\
    RCNet w/o DCN & 38.7& 23.3& 42.9& 50.8 \\
    RCNet w/o guide & 39.7& 23.7& 43.8& 52.8 \\
    RCNet w/o weight & 39.7& 23.4& 43.9& 51.8 \\
    \hline
    RCNet w/o CSN & 38.8& 20.4& 43.2& 52.8 \\
    RCNet w/o shift & 39.6& 22.8& 43.7& 53.0 \\
    RCNet w/o context & 39.4& 23.2& 43.4& 52.2 \\
  \bottomrule
\end{tabular}
\end{table}

\noindent\textbf{Comparing feature-guided upsampling with other upsampling methods.}
As shown in Table \ref{tab6}, We conduct additional experiments to compare our proposed feature-guided upsampling with other conventional upsampling operators on RCNet. It is obvious that ``$3 \times 3$ Conv'' achieves the best performance but requires a much heavier computational workload. The AP of feature-guided upsampling is only 0.1 points lower than ``$3 \times 3$ Conv'', without introducing additional overhead. As reported by Table \ref{tab6}, feature-guided upsampling yields the best efficiency, indicating that this method is able to recalibrate the high-level features before feature fusion by utilizing the location-sensitive representations for the receptive fields realignment.

\begin{table}[ht]
  \caption{Performance of various upsampling methods with RCNet using ResNet-50 backbone. Results are evaluated on COCO \textit{val}2017. ``$1 \times 1$ Conv'' (``$3 \times 3$ Conv'') indicates an extra $1 \times 1$ ($3 \times 3$) convolution layer is inserted after bilinear upsampling.
  }
  \label{tab6}
  \begin{tabular}{cccccc}
    \toprule
    Method& Params& FLOPs& AP& AP$_{50}$& AP$_{75}$\\
    \midrule
    Deconv & 39.8M& 250.9G& 39.1& 58.1& 41.8\\
    Bilinear & \textbf{37.4M}& \textbf{247.8G}& 39.7& 58.9& 42.1\\
    $1 \times 1$ Conv & 37.7M& 249.2G& 40.1& 59.4& 42.8\\
    $3 \times 3$ Conv & 39.8M& 260.3G& \textbf{40.3}& 59.4& \textbf{43.4}\\
    \hline
    FGU & \textbf{37.4M}& 247.9G& 40.2& \textbf{59.6}& 43.1\\
  \bottomrule
\end{tabular}
\end{table}

\noindent\textbf{Shift ratio of multi-scale shift module.}
In Table \ref{tab7}, we vary the shift ratio $r$ of the module and yield similar results (40.1 $ \sim $ 40.2 AP) when $r$ equals 1, 2, and 4, proving that the performance is robust to different small shift ratios. The detection performance reaches the peak by setting the shift ratio $r$ as 4. We also find RCNet earns the minimum AP increase when only 1/8 channels of feature maps are shifted. This relatively degraded performance is a natural result of deficient feature propagation.

\begin{table}[ht]
  \caption{Ablation study on the shift ratio in multi-scale shift module. The shift ratio $r$ increases from 1 to 8. Results evaluated on COCO \textit{val}2017 are reported.
  }
  \label{tab7}
  \begin{tabular}{c|cccccc}
    \toprule
    Setting& AP& AP$_{50}$& AP$_{75}$& AP$_{S}$& AP$_{M}$& AP$_{L}$ \\
    \midrule
    $r=1$& 40.1& 59.5& 42.9& 23.4& 44.0& 53.4 \\
    $r=2$& 40.1& 59.4& 42.8& \textbf{23.9}& 44.0& 53.3 \\
    $r=4$& \textbf{40.2}& \textbf{59.6}& \textbf{43.1}& 23.7& \textbf{44.2}& \textbf{53.8} \\
    $r=8$& 39.9& 59.1& 42.6& 23.6& 44.0& 53.0 \\
  \bottomrule
\end{tabular}
\end{table}

\noindent\textbf{Analysis on the effect of CSN.}
We finally delve into the necessity of CSN. CSN is designed to introduce global fusion into the feature pyramid constructed by local fusion operations. Table \ref{tab8} summarizes the detailed performance of CSN, which considerably improves the AP$_{S}$ and AR$_{S}$ by respectively 3.3 and 3.4 points compared to RCNet without CSN. The feature map $P_{l_{min}}$ with the highest resolutions only fuses adjacent high-level feature $P_{l_{min}+1}$ in RevFP that causes terrible small object detection performance. From Figure \ref{csnvis}, RCNet w/o CSN misses some small and middle objects since only a few semantics of these objects are propagated to the lowest level in RevFP. With CSN involved, semantically strong features at any other level (e.g., $P_{l_{max}}$) can be propagated to the bottom-most level $l_{min}$ via the shift operation and context modeling in CSN. 

\begin{table}[ht]
  \caption{The effect of CSN in our design. ``AR'' denotes average recall.
  }
  \label{tab8}
  \begin{tabular}{c|ccccccc}
    \toprule
    CSN& AP& AP$_{S}$& AP$_{M}$& AP$_{L}$& AR$_{S}$& AR$_{M}$& AR$_{L}$ \\
    \midrule
     & 38.8& 20.4& 43.2& 52.8& 34.0& 60.2& 72.5 \\
    \checkmark& \textbf{40.2}& \textbf{23.7}& \textbf{44.2}& \textbf{53.8}& \textbf{37.4}& \textbf{61.5}& \textbf{73.4}
    \\
  \bottomrule
\end{tabular}
\end{table}

\section{Conclusion}

In this paper, we analyze the problems of existing bidirectional feature pyramid networks in detail. To alleviate these limitations, we propose a novel architecture, named RCNet, which includes Reverse Feature Pyramid (RevFP) and Cross-scale Shift Network (CSN). RevFP is built by a single bottom-up pathway and local top-down connections to simplify the bidirectional feature fusion pipeline. CSN utilizes shift operation along scale dimension for cross-scale aggregation to enhance the correlation among all pyramid levels. The experiments on MS COCO \textit{val}2017 and \textit{test-dev} indicate that RCNet can significantly boost the detection performance of competitive baselines and achieve state-of-the-art performance with minimal computation complexity increase.

\section*{Acknowledgment}
This work is supported by the National Natural Science Foundation of China (No.61972014), and the Beijing Municipal Natural Science Foundation (No.L182014).
\bibliographystyle{ACM-Reference-Format}
\balance
\bibliography{ourbib}


\end{document}